\documentclass[11pt,a4paper]{article}
\usepackage[hyperref]{acl2021}
\usepackage{times}
\usepackage{latexsym}
\usepackage{float}
\usepackage{subcaption}

\usepackage{lipsum}
\usepackage{graphicx}

\usepackage{amssymb}
\usepackage{amsmath}

\usepackage{multirow}
\usepackage{booktabs}

\usepackage{microtype}

\usepackage{algpseudocode}
\usepackage{algorithm}

\title{Beyond Offline Mapping: \\ Learning Cross-lingual Word Embeddings through Context Anchoring}

\author{Aitor Ormazabal$^{1}$, \ Mikel Artetxe$^{2}$, \ Aitor Soroa$^{1}$, \ Gorka Labaka$^{1}$, \ Eneko Agirre$^{1}$ \\
$^1$HiTZ Center, University of the Basque Country (UPV/EHU) \\
$^2$Facebook AI Research \\
\texttt{\{aitor.ormazabal,a.soroa,gorka.labaka,e.agirre\}@ehu.eus} \\ 
\texttt{artetxe@fb.com}
}

\date{}

\aclfinalcopy %

\DeclareMathOperator*{\argmax}{arg\, max}
\DeclareMathOperator*{\csls}{CSLS}
\DeclareMathOperator*{\ctx}{ctx}

\begin{document}
\maketitle
\begin{abstract}

Recent research on cross-lingual word embeddings has been dominated by unsupervised mapping approaches that align monolingual embeddings. Such methods critically rely on those embeddings having a similar structure, but it was recently shown that the separate training in different languages causes departures from this assumption. In this paper, we propose an alternative approach that does not have this limitation, while requiring a weak seed dictionary (e.g., a list of identical words) as the only form of supervision. Rather than aligning two fixed embedding spaces, our method works by fixing the target language embeddings, and learning a new set of embeddings for the source language that are aligned with them. To that end, we use an extension of skip-gram that leverages translated context words as anchor points, and incorporates self-learning and iterative restarts to reduce the dependency on the initial dictionary. Our approach outperforms conventional mapping methods on bilingual lexicon induction, and obtains competitive results in the downstream XNLI task.

\end{abstract}

\section{Introduction}

Cross-lingual word embeddings (CLWEs) represent words from two or more languages in a shared space, so that semantically similar words in different languages are close to each other.
Early work focused on jointly learning CLWEs in two languages, relying on a strong cross-lingual supervision in the form of parallel corpora \citep{luong2015bilingual,gouws2015bilbowa} or bilingual dictionaries \citep{gouws-sogaard-2015-simple,duong2016learning}. However, these approaches were later superseded by offline mapping methods, which separately train word embeddings in different languages and align them in an unsupervised manner through self-learning  \cite{artetxe2018robust,hoshen2018nonadversarial} or adversarial training \cite{zhang2017adversarial,conneau2018word}.

Despite the advantage of not requiring any parallel resources, mapping methods critically rely on the underlying embeddings having a similar structure, which is known as the \textit{isometry assumption}. Several authors have observed that this assumption does not generally hold, severely hindering the performance of these methods \cite{sogaard2018limitations,nakashole2018characterizing,patra2019bliss}.
In later work, \citet{ormazabal-etal-2019-analyzing} showed that
this issue arises from trying to align separately trained embeddings, as joint learning methods are not susceptible to it.

In this paper, we propose an alternative approach that does not have this limitation, but can still work without any parallel resources. The core idea of our method is to fix the target language embeddings, and learn aligned embeddings for the source language from scratch. This prevents structural mismatches that result from independently training embeddings in different languages, as the learning of the source embeddings is tailored to each particular set of target embeddings. For that purpose, we use an extension of skip-gram that leverages translated context words as anchor points. So as to translate the context words, we start with a weak initial dictionary, which is iteratively improved through self-learning, and we further incorporate a restarting procedure to make our method more robust. Thanks to this, our approach can effectively work without any human-crafted bilingual resources, relying on simple heuristics (automatically generated lists of numerals or identical words) or an existing unsupervised mapping method to build the initial dictionary.
Our experiments confirm the effectiveness of our approach, outperforming previous mapping methods on bilingual dictionary induction and obtaining competitive results on zero-shot cross-lingual transfer learning on XNLI.

\section{Related work}

\label{sec:related_work}

\paragraph{Word embeddings.} Embedding methods learn static word representations based on co-occurrence statistics from a corpus. Most approaches use two different matrices to represent the words and the contexts, which are known as the \emph{input} and \emph{output} vectors, respectively \citep{mikolov2013distributed,pennington-etal-2014-glove,bojanowski2017enriching}. The output vectors play an auxiliary role, being discarded after training. Our method takes advantage of this fact, leveraging translated output vectors as anchor points to learn cross-lingual embeddings. To that end, we build on the Skip-Gram with Negative Sampling (SGNS) algorithm \citep{mikolov2013distributed}, which trains a binary classifier to distinguish whether each output word co-occurs with the given input word in the training corpus or was instead sampled from a noise distribution.

\paragraph{Mapping CLWE methods.} Offline mapping methods separately train word embeddings for each language, and then learn a mapping to align them into a shared space. Most of these methods align the embeddings through a linear map---often enforcing orthogonality constraints---and, as such, they rely on the assumption that the geometric structure of the separately learned embeddings is similar. This assumption has been shown to fail under unfavorable conditions, severely hindering the performance of these methods \citep{sogaard2018limitations,vulic-etal-2020-good}.
Existing attempts to mitigate this issue include learning non-linear maps in a latent space \citep{mohiuddin-etal-2020-lnmap}, employing maps that are only locally linear \citep{nakashole2018norma}, or learning a separate map for each word \citep{glavas-vulic-2020-non}.
However, all these methods are supervised, and have the same fundamental limitation of aligning a set of separately trained embeddings \citep{ormazabal-etal-2019-analyzing}.

\paragraph{Self-learning.} While early mapping methods relied on a bilingual dictionary to learn the alignment, this requirement was alleviated thanks to self-learning, which iteratively re-induces the dictionary during training. This enabled learning CLWEs in a semi-supervised fashion starting from a weak initial dictionary \citep{artetxe2017learning}, or in a completely unsupervised manner when combined with adversarial training \citep{conneau2018word} or initialization heuristics \citep{artetxe2018robust,hoshen2018nonadversarial}. Our proposed method also incorporates a self-learning procedure, showing that this technique can also be effective with non-mapping methods.

\paragraph{Joint CLWE methods.} Before the popularization of offline mapping, most CLWE methods extended monolingual embedding algorithms by either incorporating an explicit cross-lingual term in their learning objective, or directly replacing words with their translation equivalents in the training corpus. For that purpose, these methods relied on some form of cross-lingual supervision, ranging from bilingual dictionaries \citep{gouws-sogaard-2015-simple,duong2016learning} to parallel or document-aligned corpora \citep{luong2015bilingual,gouws2015bilbowa,vulic2016bilingual}. More recently, \citet{lample2018phrase} reported positive results learning regular word embeddings over concatenated monolingual corpora in different languages, relying on identical words as anchor points. \citet{wang2019crosslingual} further improved this approach by applying a conventional mapping method afterwards. As shown later in our experiments, our approach outperforms theirs by a large margin.

\paragraph{Freezing.} \citet{artetxe-etal-2020-cross} showed that it is possible to transfer an English transformer to a new language by freezing all the inner parameters of the network and learning a new set of embeddings for the new language through masked language modeling. This works because the frozen transformer parameters constrain the resulting representations to be aligned with English. Similarly, our proposed approach uses frozen output vectors in the target language as anchor points to learn aligned embeddings in the source language.

\section{Proposed method}
\label{sec:method}

Let $\mathbf{x}_i$ and $\mathbf{\tilde{x}}_i$ be the input and output vectors of the $i$th word in the source language, and $\mathbf{y}_j$ and $\mathbf{\tilde{y}}_j$ be their analogous in the target language.\footnote{Recall that $\{\mathbf{\tilde{x}}_i\}$ and $\{\mathbf{\tilde{y}}_j\}$ are auxiliary, and the goal is to learn aligned $\{\mathbf{x}_i\}$ and $\{\mathbf{y}_j\}$ (see \S\ref{sec:related_work}).} In addition, let $D$ be a bilingual dictionary, where $D(i) = j$ denotes that the $i$th word in the source language is translated as the $j$th word in the target language. 
Our approach first learns the target language embeddings $\{\mathbf{y}_i\}$ and $\{\mathbf{\tilde{y}}_i\}$ monolingually using regular SGNS. Having done that, we learn the source language embeddings $\{\mathbf{x}_i\}$ and $\{\mathbf{\tilde{x}}_i\}$, constraining them to be aligned with the target language embeddings according to the dictionary $D$. For that purpose, we propose an extension of SGNS that replaces the output vectors in the source language with their translation equivalents in the target language, which act as anchor points (\S \ref{subsec:anchoring}). So as to make our method more robust to a weak initial dictionary, we incorporate a self-learning procedure that re-estimates the dictionary during training (\S \ref{subsec:self_learning}), and perform iterative restarts (\S \ref{subsec:restart}). Algorithm \ref{alg:method} summarizes our method.

\begin{algorithm}[t]
\small
\begin{algorithmic}[1]
\item[\textbf{Input:} $D$ (dictionary), $C_{src}$ (src corpus), $C_{tgt}$ (tgt corpus)]
\item[\textbf{Output:} $\{\mathbf{x}_i\}$, $\{\mathbf{y}_i\}$ (aligned src and tgt embs)]
\item[\textbf{Hparams:} $T$ (updates), $R$ (restarts), $K$ (re-inductions)]
\State $\{\mathbf{y}_i\}, \{\mathbf{\tilde{y}}_i\} \gets \Call{SGNS}{C_{tgt}}$ \Comment{learn target embedings}
\For{$r \gets 1$ to  $R$} \Comment{iterative restart (\S \ref{subsec:restart})}
    \State $\{\mathbf{x}_i\}, \{\mathbf{\tilde{x}}_i\} \gets \Call{random\_init}$
    \For{$it \gets 1$ to  $T$}
        \State {$(w_i, w_j)  \gets \Call{next\_instance}{C_{src}}$}
        \State $\Call{backprop}{\mathcal{L}(w_i, w_j)}$  \Comment{core method (\S \ref{subsec:anchoring})}
        \If { $it \mod (T/K) = 0$ } \Comment{self-learn (\S \ref{subsec:self_learning})}
        \State $D \gets$ \Call{reinduce}{$\{\mathbf{x}_i\}$, $\{\mathbf{y}_i\}$}
        \EndIf
    \EndFor
\EndFor
\end{algorithmic}
\caption{Proposed method}
\label{alg:method}
\end{algorithm}

\subsection{SGNS with cross-lingual anchoring}
\label{subsec:anchoring}

Given a pair of words $(w_i, w_j)$ co-occurring in the source language corpus, we define a generalized SGNS objective as follows:
\begin{equation*}
\begin{split}
  \mathcal{L}(w_i, w_j) = \log \sigma \left( \mathbf{x}_{w_i} \cdot \ctx(w_j) \right) + \\ \sum_{i=1}^k \mathbb{E}_{w_n \sim P_n(w)} \left[ \log \sigma \left( - \mathbf{x}_{w_i} \cdot \ctx(w_n) \right) \right]
\end{split}
\label{eq:sgns}
\end{equation*}
where $k$ is the number of negative samples, $P_n(w)$ is the noise distribution, and $\ctx(w_t)$ is a function that returns the output vector to be used for $w_t$. In regular SGNS, this function would simply return the output vector of the corresponding word, so that $\ctx(w_t) = \mathbf{\tilde{x}}_{w_t}$. In contrast, our approach replaces it with its counterpart in the target language if $w_t$ is in the dictionary:
\begin{align*}
\ctx(w_t) =
\begin{cases}
\mathbf{\tilde{y}}_{D({w_t})}&\text{if $w_t \in D$} \\
\mathbf{\tilde{x}}_{w_t}&\text{otherwise}
\end{cases}
\end{align*}

During training, the replaced vectors $\{\mathbf{\tilde{y}}_i\}$ are kept frozen, acting as anchor points so that the resulting embeddings $\{\mathbf{x}_i\}$ are aligned with their counterparts $\{\mathbf{y}_i\}$ in the target language.

\subsection{Self-learning} \label{subsec:self_learning}

As shown later in our experiments, the performance of our basic method is largely dependent on the quality of the bilingual dictionary itself. However, this is not different for conventional mapping methods, which also rely on a bilingual dictionary to align separately trained embeddings in different languages. So as to overcome this issue, modern mapping approaches rely on self-learning, which alternates between aligning the embeddings and re-inducing the dictionary in an iterative fashion \citep{artetxe2017learning}.

We adopt a similar strategy, and re-induce the dictionary $D$ a total of $K$ times during training, where $K$ is a hyperparameter. To that end, we first obtain the translations for each source word using CSLS retrieval~\citep{conneau2018word}:
\begin{equation*}
D(i) = \argmax_j \csls(\mathbf{x}_i, \mathbf{y}_j)
\end{equation*}

Having done that, we discard all entries that do not satisfy the following cyclic consistency condition:\footnote{We define our cyclic consistency condition over cosine similarity, which we found to be more restrictive than CSLS (in that it discards more entries) and work better in our preliminary experiments.}
\begin{equation*}
\begin{gathered}
i \in D \iff \\
i = \argmax_k \cos \big( \mathbf{x}_k, \mathbf{y}_{\argmax_j \cos(\mathbf{x}_i, \mathbf{y}_j)} \big)
\end{gathered}
\end{equation*}

\subsection{Iterative restarts} \label{subsec:restart}

While self-learning is able to improve a weak initial dictionary throughout training, the method is still susceptible to poor local optima. This can be further exacerbated by the learning rate decay commonly used with SGNS, which makes it increasingly difficult to recover from a poor solution as training progresses. So as to overcome this issue, we sequentially run the entire SGNS training $R$ times, where $R$ is a hyperparameter of the method. We use the output from the previous run as the initial dictionary, but all the other parameters are reset and the full training process is run from scratch.

\section{Experimental setup}

We next describe the systems explored in our experiments (\S \ref{subsec:systems}), the data and procedure used to train them (\S \ref{subsec:data}), and the evaluation tasks (\S \ref{subsec:evaluation}).

\subsection{Systems} \label{subsec:systems}

We compare 3 model families in our experiments:

\paragraph{Offline mapping.} This approach learns monolingual embeddings in each of the languages separately, which are then mapped into a common space through a linear transformation. We experiment with 3 popular methods from the literature: MUSE \citep{conneau2018word}, ICP \citep{hoshen2018nonadversarial} and VecMap \citep{artetxe2018robust}. We use the original implementation of each method in their unsupervised mode with default hyperparameters.

\paragraph{Joint learning + offline mapping.} This approach jointly learns word embeddings for two languages over their concatenated monolingual corpora, where identical words act as anchor points \citep{lample2018phrase}. Having done that, the vocabulary is partitioned into one shared and two language specific subsets, which are further aligned through an offline mapping method \citep{wang2019crosslingual}. We use the joint\_align implementation from the authors with default hyperparameters, which relies on fastText for the joint learning step and MUSE for the mapping step.\footnote{The original implementation only supports the supervised mode with RCSLS mapping, so we modified it to use MUSE in the unsupervised setting as described in the original paper.}

\paragraph{Cross-lingual anchoring.} Our proposed method, described in Section \ref{sec:method}. We explore 3 alternatives to obtain the initial dictionary: \textbf{(i) identical words}, where $D_i = j$ if the $i$th source word and the $j$th target word are identically spelled, \textbf{(ii) numerals}, a subset of the former where identical words are further restricted to be sequences of digits, and \textbf{(iii) unsupervised mapping}, where we use the baseline VecMap system described above to induce the initial dictionary.\footnote{We use $\csls$ retrieval and apply the cyclic consistency restriction as described in Section \ref{subsec:self_learning}.} The first two variants make assumptions on the writing system of different languages, which is usually regarded as a weak form of supervision \citep{artetxe2017learning,sogaard2018limitations}, whereas the latter is strictly unsupervised, yet dependant on an additional system from a different family.

\subsection{Data and training details} \label{subsec:data}

\begin{table}[t]
\begin{center}
\begin{small}
  \addtolength{\tabcolsep}{-2.5pt}
  \begin{tabular}{lrrrrrrr}
    \toprule
    & en & de & es & fr & fi & ru & zh \\
    \midrule
    Tokens & 2,390 & 860 & 601 & 724 & 91 & 498 & 234 \\
    Sentences & 101 & 42 & 22 & 28 & 6 & 25 & 10 \\
    \bottomrule
\end{tabular}%
\end{small}
\end{center}
\caption{Size of the training corpora (millions).}
\label{tab:corpus_stats}
\end{table}

\begin{table}[t]
\begin{center}
\begin{small}
  \addtolength{\tabcolsep}{-2.5pt}
  \begin{tabular}{lrrrrrr}
    \toprule
    & de-en & es-en & fr-en & fi-en & ru-en & zh-en \\
    \midrule
    Identical & 44.8 & 57.6 & 63.8 & 37.7 & 4.3 & 3.3 \\
    Numerals & 1.4 & 1.6 & 1.6 & 2.4 & 1.1 & 0.2 \\
    Mapping & 53.3 & 67.3 & 69.7 & 22.3 & 28.2 & 17.1 \\
    \bottomrule
\end{tabular}%
\end{small}
\end{center}
\caption{Size of the initial dictionaries (thousands).}
\label{tab:dict_stats}
\end{table}

\begin{table*}[t]
\begin{center}
\begin{small}
  \addtolength{\tabcolsep}{-3pt}
\resizebox{\textwidth}{!}{%
  \begin{tabular}{lccccccccccccccccccc|ccc}
    \toprule
    && \multicolumn{2}{c}{de-en} && \multicolumn{2}{c}{es-en} && \multicolumn{2}{c}{fr-en} && \multicolumn{2}{c}{fi-en} && \multicolumn{2}{c}{ru-en} && \multicolumn{2}{c}{zh-en} &&& \multirow{2}{*}{avg} & \\
    \cmidrule{3-4} \cmidrule{6-7} \cmidrule{9-10} \cmidrule{12-13} \cmidrule{15-16} \cmidrule{18-19}
    && $\rightarrow$ & $\leftarrow$ && $\rightarrow$ & $\leftarrow$ && $\rightarrow$ & $\leftarrow$ && $\rightarrow$ & $\leftarrow$ && $\rightarrow$ & $\leftarrow$ && $\rightarrow$ & $\leftarrow$ &&& & \\
    \midrule
    \multicolumn{23}{c}{\textsc{Offline mapping}} \\
    \midrule
MUSE \citep{conneau2018word} && 72.9 &  74.8 &&  83.5 & 83.0 &&  81.7 & 82.3 && 0.3$^*$ & 0.3$^*$ && 0.0$^*$ & 0.3$^*$ &&  39.5 &  30.9 &&& 45.8 & \\
ICP \citep{hoshen2018nonadversarial} &&  73.9 & 75.1 &&  82.5 & 83.2 &&  80.5 & 82.3 && 0.3$^*$ & 0.3$^*$ && 59.5 &  46.1 && 0.1$^*$ & 2.8$^*$ &&& 48.9 & \\
VecMap \citep{artetxe2018robust} && 74.5 & 76.6 && 83.5 & 83.3 && 82.7 & 83.0 && 61.9 & 45.1 && \textbf{65.7} & 49.0 && 42.4 & 33.4 &&& 65.1 & \\
\midrule
\multicolumn{23}{c}{\textsc{Joint learning + offline mapping}} \\
\midrule
Joint\_Align \citep{wang2019crosslingual} &&  70.7 & 68.7 && 71.9 & 69.6 && 79.2 & 78.0 && 33.1 & 29.1 && 31.3 & 25.1 && 3.6$^*$ & 18.4 &&& 48.2 & \\
\midrule
\multicolumn{23}{c}{\textsc{Cross-lingual anchoring}} \\
\midrule
Ours (identical init) && 76.7& 77.9 &&\textbf{86.5} & 84.1 && \textbf{85.0} & 84.8 && 63.3 & 51.3 && 65.3 & \textbf{51.6} && 42.1 & \textbf{38.9} &&& 67.3 & \\
Ours (numeral init) && \textbf{76.9} & 77.7 && 86.3 & 84.1 && \textbf{85.0} & \textbf{84.9} && 63.6 & 50.6 && 64.9 & 51.4 && 1.4$^*$ & 4.9$^*$ &&& 61.0 & \\
Ours (mapping init) && 76.8 & \textbf{78.1} && 86.3 & \textbf{84.2} && 84.9 & \textbf{84.9} && \textbf{64.2} & \textbf{51.5} && \textbf{65.7} & 51.5 && \textbf{42.5} & 38.8 &&& \textbf{67.5} & \\
    \bottomrule
    \end{tabular}}
\end{small}  
\end{center}
\caption{Main BLI results on the MUSE dataset (P@1). Asterisks denote divergence ($<5\%$ P@1).}
\label{tab:bli}
\end{table*}

We learn CLWEs between English and six other languages: German, Spanish, French, Finnish, Russian and Chinese. Following common practice, we use Wikipedia as our training corpus,\footnote{We extracted the corpus from the February 2019 dump using the WikiExtractor tool.} which we preprocessed using standard Moses scripts, and restrict our vocabulary to the most frequent 200K tokens per language. In the case of Chinese, word segmentation was done using the Stanford Segmenter. Table \ref{tab:corpus_stats} summarizes the statistics of the resulting corpora, while Table \ref{tab:dict_stats} reports the sizes of the initial dictionaries derived from it for our proposed method.

For joint\_align, we directly run the official implementation over our tokenized corpus as described above. All the other systems take monolingual embeddings as input, which we learn using the SGNS implementation in word2vec.\footnote{We use 10 negative samples, a sub-sampling threshold of 1e-5, 300 dimensions, and 10 epochs. Note that joint\_align also learns 300-dimensional vectors, but runs fastText with default hyperparameters under the hood.} For our proposed method, we set English as the target language, fix the corresponding monolingual embeddings, and learn aligned embeddings in the source language using our extension of SGNS (\S \ref{sec:method}).\footnote{In our preliminary experiments, we observed our proposed method to be quite sensitive to which language is the source and which one is the target. We find the language with the largest corpus to perform best as the target, presumably because the corresponding monolingual embeddings are better estimated, so it is more appropriate to fix them and learn aligned embeddings for the other language. Following this observation, we set English as the target language for all pairs, as it is the language with the largest corpus.}
We set the number of restarts $R$ to 3, the number of reinductions per restart $K$ to 50, and the number of epochs to $10\frac{\text{\#trg sents}}{\text{\#src sents}}$, which makes sure that the source language gets a similar number of updates to the 10 epochs done for English.\footnote{For a fair comparison, we also tried using the same number of epochs for the baseline systems, but this performed worse than the reported numbers with 10 epochs.} For all the other hyperparameters, we use the same values as for the monolingual embeddings.  We made all of our development decisions based on preliminary experiments on English-Finnish, without any systematic hyperparameter exploration. 
Our implementation runs on CPU, except for the dictionary reinduction steps, which run on a single GPU for around one hour in total.

\subsection{Evaluation tasks} \label{subsec:evaluation}

As described next, we evaluate our method on two tasks: Bilingual Lexicon Induction (BLI) and Cross-lingual Natural Language Inference (XNLI).

\paragraph{BLI.} Following common practice, we induce a bilingual dictionary through $\csls$ retrieval \citep{conneau2018word} for each set of cross-lingual embeddings, and evaluate the precision at 1 (P@1) with respect to the gold standard test dictionary from the MUSE dataset \citep{conneau2018word}. For the few out-of-vocabulary source words, we revert to copying as a back-off strategy,\footnote{This has a negligible impact in practice, as it accounts for less than 1.4\% of the test cases. Moreover, all of our systems use the same underlying vocabulary, so they are affected in the exact same way.} so our reported numbers are directly comparable to prior work in terms of coverage.

\paragraph{XNLI.} We train an English natural language inference model on MultiNLI \citep{williams-etal-2018-broad}, and evaluate the zero-shot cross-lingual transfer performance on the XNLI test set \citep{conneau-etal-2018-xnli} for the subset of our languages covered by it. To that end, we follow \citet{glavas-etal-2019-properly} and train an Enhanced Sequential Inference Model (ESIM) on top of our original English embeddings, which are kept frozen during training. At test time, we transfer into the rest of the languages by plugging in the corresponding aligned embeddings. Note that we use the exact same English model for our proposed method and the baseline MUSE and ICP systems,\footnote{This is possible because they all fix the target language embeddings (English in this case) and align the embeddings in the source language with them, either through mapping (MUSE, ICP) or learning from scratch (ours).} which only differ in the set of aligned embeddings used for cross-lingual transfer. In contrast, VecMap and joint\_align also manipulate the target English embeddings, which would require training a separate model for each language pair, so we decide to exclude them from this set of experiments.\footnote{In addition to the computational overhead, having separate models introduces some variance, while our comparison is more direct.}

\begin{table*}[t]
\begin{center}
\begin{small}
  \addtolength{\tabcolsep}{-2.5pt}
  \begin{tabular}{lccccccccccccc|ccc}
    \toprule
    && \multicolumn{2}{c}{de-en} && \multicolumn{2}{c}{es-en} && \multicolumn{2}{c}{fr-en} && \multicolumn{2}{c}{ru-en} &&& \multirow{2}{*}{avg} & \\
    \cmidrule{3-4} \cmidrule{6-7} \cmidrule{9-10} \cmidrule{12-13}
    && $\rightarrow$ & $\leftarrow$ && $\rightarrow$ & $\leftarrow$ && $\rightarrow$ & $\leftarrow$ && $\rightarrow$ & $\leftarrow$ &&& & \\

    \midrule
    
     \citet{conneau2018word}          && 72.2  & 74.0 && 83.3  & 81.7  && 82.1  & 82.3  && 59.1  & 44.0  &&& 72.3 & \\
     \citet{hoshen2018nonadversarial} && 73.0  & 74.7 && 84.1  & 82.1  && 82.9  & 82.3  && 61.8  & 47.5  &&& 73.6 & \\
     \citet{grave2018unsupervised}    && 73.3  & 75.4 && 84.1  & 82.8  && 82.9  & 82.6  && 59.1  & 43.7  &&& 73.0 & \\
     \citet{alvarezmelis2018gromov}   && 72.8  & 71.9 && 80.4  & 81.7  && 78.9  & 81.3  && 43.7  & 45.1  &&& 69.5 & \\
     \citet{yang2018learning}         && 70.3  & 71.5 && 79.3  & 79.9  && 78.9  & 78.4  && -     & -     &&& -    & \\
     \citet{mukherjee2018learning}    && -     & -    && 79.2  & \textbf{84.5}  && -     & -     && -     & -     &&& -  &  \\
     \citet{alvarez2018towards}       && 71.1  & 73.8 && 81.8  & 81.3  && 81.6  & 82.9  && 55.4  & 41.7  &&& 71.2 & \\
     \citet{xu2018unsupervised}       && 67.0  & 69.3 && 77.8  & 79.5  && 75.5  & 77.9  && -     & -     &&& - &  \\
     \citet{wang2019crosslingual}     && 72.2  & 74.2 && 84.2  & 81.4  && 83.6  & 82.8  && 58.3  & 45.0  &&& 72.7 & \\
     \citet{zhou-etal-2019-density}  && 74.4  & 77.2 && 84.9  & 82.8  && 83.5  & 83.1  && 63.6    & 49.2     &&& 74.8   &  \\
     \citet{li-etal-2020-simple}      && 74.3  & 75.3 && 84.6  & 82.4  && 83.7  & 82.6  && -     & -     &&& -    &  \\
    \midrule 

    Ours (mapping init)      && \textbf{76.8} & \textbf{78.1} && \textbf{86.3} & 84.2 && \textbf{84.9} & \textbf{84.9} && \textbf{65.7} & \textbf{51.5} &&& \textbf{76.6} & \\

    \bottomrule
  \end{tabular}%
\end{small}
\end{center}
\caption{BLI results on MUSE dataset in comparison with prior published results (P@1). All systems are fully unsupervised (except that of \citet{zhou-etal-2019-density}, which uses identical words as a seed dictionary), and use SGNS embeddings trained on Wikipedia.} 
\label{tab:sota}
\end{table*}

\section{Results} 
\label{sec:results}

We next discuss our main results on BLI (\S\ref{subsec:main}) and XNLI (\S\ref{subsec:xnli}), followed by our ablation study (\S \ref{subsec:ablation}) and error analysis (\S \ref{subsec:qualitative}) on BLI.

\subsection{BLI} \label{subsec:main}

Table \ref{tab:bli} comprises our main BLI results. We observe that our method obtains the best results in all directions (matched by VecMap in Russian-English), outperforming the strongest baseline by 2.4 points on average for the mapping based initialization. Our improvements are more pronounced in the backward direction (3.1 points on average) but still substantial in the forward direction (1.7 points on average).

It is worth noting that some systems fail to converge to a good solution for the most challenging language pairs. This includes our proposed method in the case of Chinese-English when using the numeral-based initialization, which we attribute to the smaller size of the initial dictionary (only 244 entries, see Table \ref{tab:dict_stats}). Other than that, we observe that our approach obtains very similar results regardless of the initial dictionary. Quite remarkably, the variant using VecMap for initialization (\textit{mapping init}) is substantially stronger than VecMap itself despite not using any additional training signal.

So as to put our results into perspective, Table \ref{tab:sota} compares them to previous numbers reported in the literature. Note that the numbers are comparable in terms of coverage and all systems use Wikipedia as the training corpus, although they might differ on the specific dump used and the preprocessing details.\footnote{In particular, most mapping methods use the official Wikipedia embeddings from fastText. Unfortunately, the pre-processed corpus used to train these embeddings is not public, so works that explore other approaches, like ours, need to use their own pre-processed copy of Wikipedia.} As it can be seen, our approach obtains the best results by a substantial margin.\footnote{\citet{artetxe-etal-2019-bilingual} report even stronger results based on unsupervised machine translation instead of direct retrieval with CLWEs. Note, however, that their method still relies on cross-lingual embeddings to build the underlying phrase-table, so our improvements should be largely orthogonal to theirs.}

\subsection{XNLI} \label{subsec:xnli}

\begin{table}[t]
\small
\begin{center}
  \addtolength{\tabcolsep}{-2.5pt}
  \resizebox{\columnwidth}{!}{%
  \begin{tabular}{lllllll}
    \toprule
     & \multicolumn{1}{c}{en} & \multicolumn{1}{c}{de} & \multicolumn{1}{c}{es} & \multicolumn{1}{c}{fr} & \multicolumn{1}{c}{ru} & \multicolumn{1}{c}{zh} \\
    \midrule
MUSE & \textbf{73.9} & 65.0 & 68.1 &\textbf{67.9} & 39.1$^*$ & \textbf{61.4}   \\
ICP & \textbf{73.9} & 62.2 & 64.2 & 65.7 & 59.4 & 36.1$^*$  \\
\midrule
Ours (identical init) & \textbf{73.9} & 65.0 &\textbf{68.7} & 67.1 & \textbf{63.5} & 49.8 \\
Ours (numeral init) & \textbf{73.9} & 65.0 & 68.6 & 67.1 & 63.3 & 34.9$^*$ \\
Ours (mapping init) & \textbf{73.9} & \textbf{65.1} & 68.6 & 67.0 & \textbf{63.5} & 49.4   \\

    \bottomrule
    \end{tabular}}
\end{center}
\caption{XNLI results (accuracy). Asterisks denote divergence ($<5\%$ P@1 in BLI).}
\label{tab:xnli}
\end{table}

We report our XNLI results in Table \ref{tab:xnli}. We observe that our method is competitive with the baseline mapping systems, achieving the best results on 3 out of the 5 transfer languages by a small margin. Nevertheless, it significantly lags behind MUSE on Chinese, even if the exact same set of cross-lingual embeddings performed better than MUSE at BLI. While striking, similar discrepancies between BLI and XNLI performance where also observed in previous studies \citep{glavas-etal-2019-properly}. Finally, we observe that the initial dictionary has a negligible impact in the performance of our proposed method, which supports the idea that our approach converges to a similar solution given any reasonable initialization.

\subsection{Ablation study} \label{subsec:ablation}

\begin{table}[t]
\begin{center}
\begin{small}
  \begin{tabular}{lr}
    \toprule
Basic method (identical init) & 53.9 \\
\quad + \textit{self-learning} & 66.9 \\
\quad \quad + \textit{iterative restarts} & 67.3 \\
\midrule
Basic method (numeral init) & 2.6 \\
\quad + \textit{self-learning} & 53.9 \\
\quad \quad + \textit{iterative restarts} & 61.0 \\
\midrule
Basic method (mapping init) & 67.5 \\
\quad + \textit{self-learning} & 67.5 \\
\quad \quad + \textit{iterative restarts} & 67.5 \\
\bottomrule
\end{tabular}
\end{small}  
\end{center}
\caption{Ablation results on BLI (average P@1)}
\label{tab:ablation}
\end{table}

\begin{figure*}[ht]
\centering

  \includegraphics[clip,width=\linewidth]{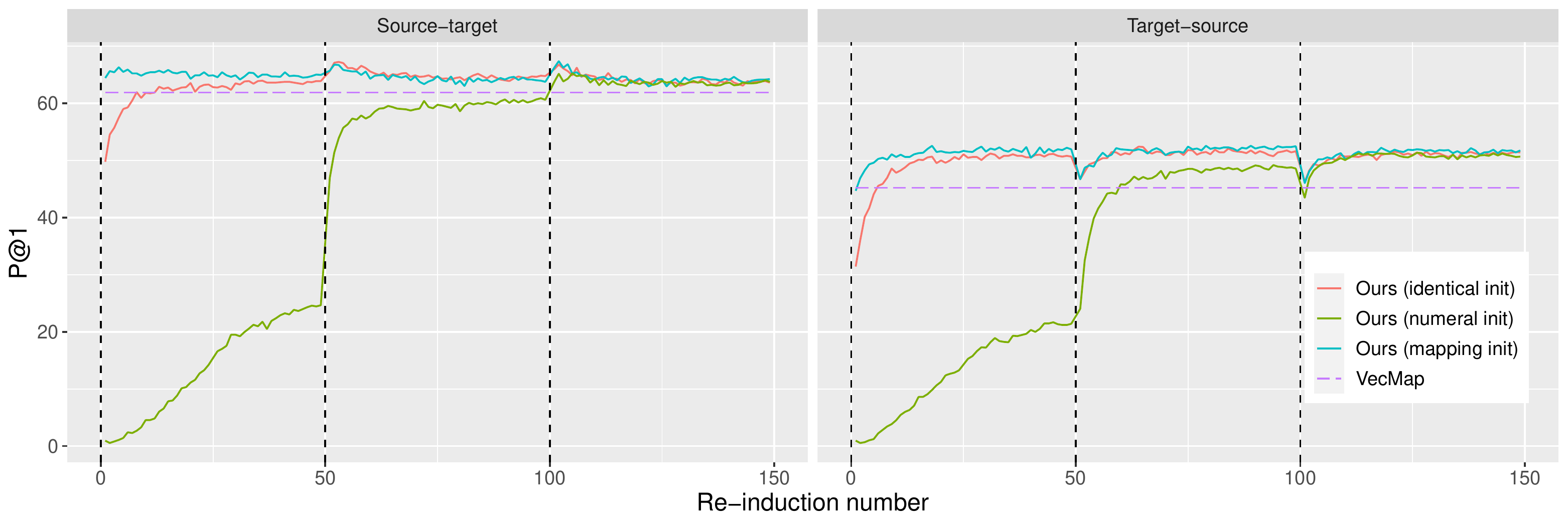}%

\caption{Finnish-English learning curves (BLI P@1). The iterative restarts happen at the vertical lines.}

\label{fig:learncurve}
\end{figure*}

So as to understand the role of self-learning and the iterative restarts in our approach, we perform an ablation study and report our results in Table \ref{tab:ablation}. We observe that the contribution of these components is greatly dependant on the initial dictionary. For the numeral initialization, the basic method works poorly, and both extensions bring large improvements. In contrast, the identical initialization does not benefit from iterative restarts, but self-learning still plays a major role. In the case of the mapping-based initialization, the basic method is already very competitive. This suggests that both the self-learning and the iterative restarts are helpful to make the method more robust to a weak initialization, and have a minor impact otherwise.

In order to better understand the underlying learning dynamics, we analyze the learning curves for Finnish-English in Figure \ref{fig:learncurve}. We observe that, when the initial dictionary is strong, our method surpasses the baseline and stabilizes early. In contrast, convergence is much slower when using the weak numeral-based initialization, and the iterative restarts are critical to escape poor local optima.

\begin{table*}[t]
\begin{center}
\begin{small}
  \begin{tabular}{ccccccccc}
    \toprule
    &&&&&& \multicolumn{3}{c}{Examples} \\
    \cmidrule{7-9}
    Gold standard && Type && Cases && Source & VecMap & Ours  \\
    \midrule
    \multirow{9}{*}{\shortstack{Ours right\\ -- \\ VecMap wrong}}

    && \multirow{2}{*}{Common errors} && \multirow{2}{*}{30.5\%} && derrotas & victories & defeats \\
    &&&&&& campeona & medalist & champion \\
    \cmidrule{3-9}
    && \multirow{2}{*}{Named entity, ours copies} && \multirow{2}{*}{21.1\%} && philadelphia & pittsburgh & philadelphia \\
    &&&&&& susana & beatriz & susana \\
    \cmidrule{3-9}
    && \multirow{2}{*}{EN word in ES vocab, ours copies} && \multirow{2}{*}{15.8\%} && pink & tangerine & pink \\
    &&&&&& space & sci & space \\
    \cmidrule{3-9}
    && \multirow{2}{*}{Gap in gold standard} && \multirow{2}{*}{5.3\%} && adecuada & appropriate & adequate \\
    &&&&&& marquesa & marchioness & marquise \\

    \midrule
    \multirow{7}{*}{\shortstack{VecMap right\\ -- \\ Ours wrong}}
    && \multirow{2}{*}{Common errors} && \multirow{2}{*}{15.8\%} && conservadores & conservatives & liberals \\
    &&&&&& noveno & ninth & tenth \\
    \cmidrule{3-9}
    && \multirow{2}{*}{ES word in EN vocab, ours copies} && \multirow{2}{*}{7.4\%} && calzada & roadway & calzada \\
    &&&&&& cantera & quarry & cantera \\
    \cmidrule{3-9}
    && \multirow{2}{*}{Gap in gold standard} && \multirow{2}{*}{4.2\%} && ferroviario & railway & rail \\
    &&&&&& situados & situated & positioned \\
\bottomrule
\end{tabular}
\end{small}  
\end{center}
\caption{BLI error analysis on Spanish-English. See Section \ref{subsec:qualitative} for details. }
\label{tab:qualitative}
\end{table*}

\begin{table*}[t]
\begin{center}
\begin{small}
  \addtolength{\tabcolsep}{-3pt}
\resizebox{\textwidth}{!}{%
  \begin{tabular}{lccccccccccccccccccc|ccc}
    \toprule
    && \multicolumn{2}{c}{de-en} && \multicolumn{2}{c}{es-en} && \multicolumn{2}{c}{fr-en} && \multicolumn{2}{c}{fi-en} && \multicolumn{2}{c}{ru-en} && \multicolumn{2}{c}{zh-en} &&& \multirow{2}{*}{avg} & \\
    \cmidrule{3-4} \cmidrule{6-7} \cmidrule{9-10} \cmidrule{12-13} \cmidrule{15-16} \cmidrule{18-19}
    && $\rightarrow$ & $\leftarrow$ && $\rightarrow$ & $\leftarrow$ && $\rightarrow$ & $\leftarrow$ && $\rightarrow$ & $\leftarrow$ && $\rightarrow$ & $\leftarrow$ && $\rightarrow$ & $\leftarrow$ &&& & \\
    \midrule

VecMap \citep{artetxe2018robust} && 68.3 & 70.2 && 85.1 & 79.4 && 80.8 & 78.1 && \textbf{58.4} & 38.9 && \textbf{66.1} & 48.6 && 45.0 & 34.5 &&& 62.8 & \\

Joint\_Align \citep{wang2019crosslingual} &&  57.0 & 53.3 && 63.0 & 57.4 && 70.2 & 64.4 && 4.0$^*$ & 0.7$^*$ && 31.3 & 22.4 && 3.5$^*$ & 0.9$^*$ &&& 35.7 & \\

\midrule

Ours (identical init) && 68.9 & 72.2 && \textbf{86.0} & 80.7 && 81.5 & 80.0 && 54.0 & 41.0 && 65.7 & 50.9 && 44.6 & \textbf{38.1} &&& 63.6 & \\

Ours (mapping init) && \textbf{68.9} & \textbf{72.3} &&        {85.8} & \textbf{80.8} && \textbf{81.4} & \textbf{80.2} && 55.4 & \textbf{41.6} && \textbf{66.1} & \textbf{51.0} && \textbf{45.1} &        {37.9} &&& \textbf{63.9} & \\
    \bottomrule
    \end{tabular}}
\end{small}  
\end{center}
\caption{BLI results on MUSE with identical words removed (P@1). Asterisks denote divergence ($<5\%$ P@1).}
\label{tab:bli_nosame}
\end{table*}

\subsection{Error analysis}
\label{subsec:qualitative}

So as to better understand where our improvements in BLI are coming from, we perform an error analysis on the Spanish-English direction. To that end, we manually inspect the 69 instances for which our method (with mapping-based initialization) produced a correct translation while VecMap failed according to the gold standard, as well as the 26 instances for which the opposite was true. We then categorize these errors into several types, which are summarized in Table \ref{tab:qualitative}.

We observe that, in 52.6\% of the 95 analyzed instances, the translation produced by our method is identical to the source word, while this percentage goes down to 4.2\% for VecMap. This tendency of our approach to copy its input is striking, as the model has no notion about the words being identically spelled.\footnote{The variants of our system with identical or numeral initialization do indirectly see this signal, but the one analyzed here is initialized with the VecMap mapping.} A large portion of these cases correspond to named entities, where copying is the right behavior, while VecMap outputs a different proper noun. There are also some instances where the input word is in the target language,\footnote{English words will often appear in other languages as part of named entities (e.g., ``pink'' as part of ``Pink Floyd''), which explains the presence of such words in the Spanish vocabulary.} which can be considered an artifact of the dataset, but copying also seems the most reasonable behavior in these cases.
Finally, there are also a few cases where the input word is present in the target vocabulary, which is selected by our method and counted as an error. Once again, we consider these to be an artifact of the dataset, as copying seems a reasonable choice if the input word is considered to be part of the target language vocabulary. The remaining cases where neither method copies mostly correspond to common errors, where one of the systems (most often VecMap) outputs a semantically related but incorrect translation. However, there are also a few instances where both translations are correct, but one of them is missing in the gold standard.

With the aim to understand the impact of identical words in our original results, we re-evaluated the systems using a filtered version of the MUSE gold standard dictionaries, where we removed all source words that were included in the set of candidate translations. In order to be fair, we filtered out identical words from the output of the system, reverting to the second highest-ranked translation whenever the first one is identical to the source word. The results for the strongest system in each family are shown in Table \ref{tab:bli_nosame}. Even if the margin of improvement is reduced compared to Table \ref{tab:bli}, the best results are still obtained by our proposed method,
bringing an average improvement of 1.1 points. It is also worth noting that joint\_align, which shares a portion of the vocabulary for both languages (and will thus translate all words in the shared vocabulary identically), suffers a large drop in performance. This highlights the importance of accompanying quantitative BLI evaluation with an error analysis as urged by previous studies \citep{kementchedjhieva2019lost}.

\section{Conclusions and future work} 
\label{sec:conclusions}

Our approach for learning CLWEs addresses the main limitations of both offline mapping and joint learning methods. Different from mapping approaches, it does not suffer from structural mismatches arising from independently training embeddings in different languages, as it works by constraining the learning of the source embeddings so they are aligned with the target ones. At the same time, unlike previous joint methods, our system can work without any parallel resources, relying on numerals, identical words or an existing mapping method for the initialization. We achieve this by combining cross-lingual anchoring with self-learning and iterative restarts. While recent research on CLWEs has been dominated by mapping approaches, our work shows that the fundamental techniques that popularized these methods (e.g., the use of self-learning to relax the need for cross-lingual supervision) can also be effective beyond this paradigm.

Despite its simplicity, our experiments on BLI show the superiority of our method when compared to previous mapping systems. We complement these results with additional experiments on a downstream task, where our method obtains competitive results, as well as an ablation study and a systematic error analysis. We identify a striking tendency of our method to translate words identically, even if it has no notion of the words being identically spelled. Thanks to this, our method is particularly strong at translating named entities, but we show that our improvements are not limited to this phenomenon. These insights confirm the value of accompanying quantitative results on BLI with qualitative evaluation \citep{kementchedjhieva2019lost} and/or other tasks \citep{glavas-etal-2019-properly}. 

In the future, we would like to further explore CLWE methods that go beyond the currently dominant mapping paradigm. In particular, we would like to remove the requirement of a seed dictionary altogether by using adversarial learning, and explore more elaborated context translation and dictionary re-induction schemes. 

\section*{Acknowledgments}

Aitor Ormazabal, Aitor Soroa, Gorka Labaka and Eneko Agirre were
supported by the Basque Government (excellence research group IT1343-19
and DeepText project KK-2020/00088), project BigKnowledge
(\textit{Ayudas Fundación BBVA a equipos de investigación científica 2018}) and the Spanish MINECO (project DOMINO PGC2018-102041-B-I00
MCIU/AEI/FEDER, UE). Aitor Ormazabal was supported by a doctoral grant
from the Spanish MECD.

\bibliography{acl2021}
\bibliographystyle{acl_natbib}

\end{document}